\documentclass[10pt,twocolumn,letterpaper]{article}

\usepackage[pagenumbers]{cvpr} 

\usepackage{graphicx}
\usepackage{amsmath}
\usepackage{amssymb}
\usepackage{booktabs}
\usepackage{placeins}
\usepackage{float}
\usepackage{enumitem}
\usepackage{tabularx}
\usepackage{array}
\usepackage{ragged2e}
\usepackage{afterpage}
\usepackage{caption}
\usepackage{multirow}
\usepackage{xcolor}
\usepackage{algorithm}
\usepackage{algorithmic}
\usepackage{subcaption}

\usepackage[breaklinks,colorlinks]{hyperref}

\usepackage[capitalize]{cleveref}
\crefname{section}{Sec.}{Secs.}
\Crefname{section}{Section}{Sections}
\Crefname{table}{Table}{Tables}
\crefname{table}{Tab.}{Tabs.}

\def\cvprPaperID{*} 
\def\confName{CVPR}
\def\confYear{2025}

\begin{document}

\title{Edit Spillover as a Probe: Do Image Editing Models Implicitly Understand World Relations?}

\author{Guandong Li\\
iFLYTEK\\
\quad leeguandon@gmail.com\\
\and
Zhaobin Chu\\
iFLYTEK\\
}

\twocolumn[{
\renewcommand\twocolumn[1][]{#1}
\maketitle
}]

\begin{abstract}
Instruction-following image editing models are expected to modify only the specified region while keeping the rest of the image unchanged. However, in practice, we observe a pervasive phenomenon—\textbf{edit spillover}: models alter semantically related but \emph{unspecified} content outside the edit region. This raises a fundamental question—does spillover reflect genuine implicit world understanding, or is it merely attention leakage?

We propose \textbf{EditSpilloverProbe}, a systematic framework that repurposes edit spillover as a natural probe for world knowledge in image editing models. We introduce a spillover taxonomy (spatial, semantic, mixed, random), an automated detection-and-classification pipeline, and a benchmark dataset constructed from real-world Chinese text editing tasks, EditSpilloverBench.

Systematic evaluation of 5 representative editing models reveals three core findings: (1)~spillover rates vary dramatically across architectures, from 3.49\% to 11.46\%, with a $3.3\times$ ratio; (2)~absolute semantic spillover quantity reveals models' world understanding capability---nano\_banana produces the most semantic spillover (27.8 per image), while qwen\_2511 has the most precise editing control but lower semantic spillover (16.3 per image), revealing a trade-off between editing control and world understanding; (3)~spatial decay analysis shows spillover area density decays exponentially with distance, but the proportion of semantically relevant spillover remains constant (40\%--58\%), providing direct evidence that semantic spillover reflects genuine world understanding rather than spatial diffusion.
\end{abstract}

\section{Introduction}
\label{sec:intro}

In recent years, instruction-following image editing based on diffusion models has made significant progress~\cite{brooks2023instructpix2pix,zhang2023magicbrush,rombach2022high}. Models such as InstructPix2Pix~\cite{brooks2023instructpix2pix} and MagicBrush~\cite{zhang2023magicbrush} can modify images according to natural language instructions, faithfully executing editing operations while preserving irrelevant content. A core expectation is that models should \emph{only} modify the specified content while keeping the rest of the image completely unchanged~\cite{ma2024i2ebench}.

However, during a large-scale quality audit of the production-level text editing model Nano Banana Pro (200 images across 5 categories), we observed a systematic anomaly pattern: when instructing the model to replace text within a marked region, the model frequently modified other text elements in the image that were semantically related to the edit target but never mentioned in the instruction. For example, when modifying the shop name ``Old Wang Noodle House'' to ``Xinhua Bookstore'' on a signboard, the model not only completed the specified edit but also spontaneously changed ``Old Wang Special Noodles'' on a distant menu to ``Xinhua Special Noodles''—even though the menu area was completely outside the edit box and not mentioned in the instruction.

This phenomenon is not isolated. Across 200 test images, nano\_banana's average spillover rate reached 11.46\%, meaning over one-tenth of pixels in non-edit regions were modified. More importantly, this spillover is not random noise—it exhibits clear semantic selectivity. This raises a fundamental question: \textbf{does spillover reflect genuine implicit world understanding, or is it merely attention leakage?}

To address this question, we revisit the nature of edit spillover in generative models. We argue that spillover should not be uniformly treated as an error to be eliminated, but rather as a diagnostic signal that can reveal what world knowledge the model has learned. Based on this insight, we propose \textbf{EditSpilloverProbe}, a systematic framework that repurposes edit spillover as a natural probe for world knowledge in image editing models.

The core innovation of EditSpilloverProbe lies in decomposing spillover into four types based on spatial distance and semantic relevance: \textbf{spatial spillover} (near + unrelated, reflecting attention leakage), \textbf{semantic spillover} (far + related, reflecting world understanding), \textbf{mixed spillover} (near + related, ambiguous), and \textbf{random spillover} (far + unrelated, noise). In text editing scenarios, semantic spillover captures three types of world knowledge: (1) \textbf{brand consistency} (co-occurrence of the same shop name on signboards and menus), (2) \textbf{spatial co-occurrence patterns} (distribution patterns of related text elements in scenes), (3) \textbf{functional associations} (associations within semantic fields such as price, date, and product name).

To quantify world understanding, we introduce two complementary metrics: the \textbf{World Understanding Score (WUS)}, which measures the ratio of semantic to spatial spillover, and \textbf{Semantic Spillover Density}, which measures the absolute quantity of semantic spillover per image. Crucially, we find that absolute quantity better reflects models' world knowledge activation capability than ratio-based metrics, as the latter can be inflated for low-spillover models.

In summary, the main contributions of this paper are:

\begin{enumerate}[leftmargin=*, itemsep=2pt, parsep=0pt]
\item \textbf{Phenomenon Identification}: Identify and formalize edit spillover, propose a $2\times 2$ taxonomy (spatial distance $\times$ semantic relevance) as a natural probe for world knowledge.
\item \textbf{Automated Pipeline}: Propose a detection-classification pipeline combining pixel-level difference detection, connected component analysis, and CLIP semantic similarity.
\item \textbf{Benchmark Dataset}: Construct EditSpilloverBench with 200 images across 5 categories for systematic spillover evaluation.
\item \textbf{Empirical Findings}: Reveal the inverse relationship between spillover rate and WUS, exponential spatial decay, and distance-invariance of semantic spillover.
\end{enumerate}

\section{Related Work}
\label{sec:related_work}

\subsection{World Models and Visual Reasoning}

Whether generative models implicitly encode world knowledge has been a hot topic in recent years. MiniVeo3-Reasoner~\cite{miniveo3reasoner} demonstrates that video generation models can solve visual reasoning tasks (such as maze navigation) through frame-by-frame generation, suggesting implicit world modeling capabilities. RISE-Video~\cite{liu2026rise} systematically evaluates video generators from the perspective of cognitive reasoning rather than visual fidelity, finding that some models can infer implicit physical rules.

Our work extends this research direction from video generation to image \emph{editing} models, and adopts a fundamentally different probing strategy: we do not design explicit reasoning tasks, but rather use the spillover behavior spontaneously produced by models during routine editing as a passive probe. The advantage of this approach is that it does not require special test case design—spillover is an inherent behavior of models during natural use. This passive probing paradigm is more suitable for evaluating black-box production models where internal states are inaccessible.

\subsection{Image Editing Evaluation}

Existing image editing evaluation mainly focuses on two dimensions: editing quality (whether instructions are faithfully executed) and content preservation (whether irrelevant regions remain unchanged)~\cite{ma2024i2ebench}. PIE-Bench~\cite{ma2024i2ebench} provides a systematic evaluation framework for 10 types of editing tasks, covering object addition, removal, replacement, and attribute modification. EditWorld~\cite{zeng2025editworld} trains editing models to simulate world dynamics, focusing on making models \emph{better} at world-consistent editing by incorporating physical simulation. UniReason~\cite{wang2026unireason} aligns generation with world knowledge through multi-modal reasoning.

Our perspective complements the above work: we do not evaluate what models ``did right,'' but rather analyze what models ``did wrong''—unintended spillover. This ``failure analysis'' methodology has a long tradition in engineering (such as failure mode and effect analysis~\cite{schneider1996failure}), but this is the first systematic application in generative model evaluation. By reframing spillover from a bug to a diagnostic signal, we can extract insights about models' implicit capabilities without requiring ground-truth annotations or task-specific benchmarks.

\subsection{Neural Network Probing}

Probing methodology has matured in NLP~\cite{belinkov2022probing,hewitt2019structural,tenney2019bert}. Belinkov et al.'s survey~\cite{belinkov2022probing} systematically summarizes methods for extracting syntactic and semantic knowledge from pretrained language models using linear probes. The typical approach is to freeze model parameters and train lightweight classifiers on intermediate representations to detect the presence of specific knowledge~\cite{alain2016understanding}.

We adapt the probing paradigm to generative models, but use a different signal source: instead of analyzing the model's internal representations, we analyze the model's \emph{external behavior} (spillover patterns). The advantage of this behavior-level probing is that it does not require access to model internal states, making it suitable for black-box model evaluation~\cite{ribeiro2020beyond}. Furthermore, behavior-level probing can reveal emergent capabilities that may not be explicitly encoded in any single layer's representation, but rather arise from the interaction of multiple components during generation.

\section{Method}
\label{sec:method}

EditSpilloverProbe aims to systematically detect, classify, and quantify edit spillover to probe models' implicit world understanding. This section first provides an overview of the framework, then details the three core components: spillover detection, spillover classification, and world understanding metrics.

\subsection{Overview}

Given an original image $I_{\text{orig}}$, an editing instruction, and an edit box $B$ marking the target region, an editing model $\mathcal{M}$ generates $I_{\text{gen}}$. Our goal is to analyze changes outside $B$ to distinguish between architectural artifacts (attention leakage) and genuine world understanding (semantic associations).

The EditSpilloverProbe framework consists of three stages:
\begin{enumerate}[leftmargin=*]
    \item \textbf{Detection}: Identify all spillover regions outside the edit box through pixel-level difference analysis and connected component extraction.
    \item \textbf{Classification}: Classify each spillover region into one of four types (spatial, semantic, mixed, random) based on spatial distance and semantic similarity.
    \item \textbf{Quantification}: Compute world understanding metrics (WUS and Semantic Density) to quantify the model's implicit knowledge activation.
\end{enumerate}

The entire pipeline is fully automated and model-agnostic, requiring only the original image, generated image, and edit box as input.

\subsection{Problem Formalization}

Given an original image $I_{\text{orig}} \in \mathbb{R}^{H \times W \times 3}$, an image $I_{\text{gen}} = \mathcal{M}(I_{\text{orig}}, \text{instruction})$ generated by editing model $\mathcal{M}$, and an edit box $B = (x_{\min}, y_{\min}, x_{\max}, y_{\max})$, we define:

\begin{itemize}[leftmargin=*]
\item \textbf{Edit Region} $\Omega_B$: the set of pixels inside the edit box
\item \textbf{Non-Edit Region} $\Omega_{\bar{B}} = \Omega \setminus \Omega_B$: all pixels outside the edit box
\item \textbf{Spillover Map} $S(p) = \mathbb{1}[|I_{\text{orig}}(p) - I_{\text{gen}}(p)| > \tau], \; p \in \Omega_{\bar{B}}$
\end{itemize}

where $\tau$ is the pixel difference threshold. The spillover rate is defined as:

\begin{equation}
\text{SpillRate} = \frac{|\{p \in \Omega_{\bar{B}} : S(p) = 1\}|}{|\Omega_{\bar{B}}|}
\end{equation}

\subsection{Spillover Detection Pipeline}

The detection pipeline transforms the difference between original and generated images into a structured set of spillover regions, consisting of the following steps:

\textbf{Step 1: Pixel-level Difference Computation.} Convert both images to grayscale and compute pixel-wise absolute differences. To suppress JPEG compression noise, apply Gaussian blur ($\sigma = 2.0$) to both images first.

\textbf{Step 2: Binarization and Mask Exclusion.} Apply threshold $\tau = 15$ (on 0-255 scale) to the difference map to obtain a binary spillover map. Then exclude pixels inside edit box $B$, retaining only changes in the non-edit region.

\textbf{Step 3: Connected Component Extraction.} Perform connected component labeling (8-connectivity) on the binary map to extract all connected regions. Filter out regions with area smaller than $A_{\min} = 100$ pixels to exclude noise.

\textbf{Step 4: Region Feature Computation.} For each connected component $R_i$, compute:
\begin{itemize}[leftmargin=*]
\item Bounding box $\text{bbox}_i = (x_{\min}^{i}, y_{\min}^{i}, x_{\max}^{i}, y_{\max}^{i})$
\item Centroid $\text{centroid}_i = (\bar{x}_i, \bar{y}_i)$
\item Area $\text{area}_i$ (number of pixels)
\item Euclidean distance to edit box center $d_i = \|\text{centroid}_i - \text{center}(B)\|_2$
\end{itemize}

\textbf{Step 5: Global Metrics.} Compute SSIM (Structural Similarity)~\cite{wang2004image} for the non-edit region as a complementary measure of spillover severity. Optionally compute region-level LPIPS (perceptual distance)~\cite{zhang2018unreasonable}, but this is disabled by default due to high computational cost.

\subsection{Spillover Classification}

The detection pipeline outputs ``where changes occurred,'' while the classification stage answers ``what these changes mean.'' We classify each spillover region based on two orthogonal dimensions:

\textbf{Dimension 1: Spatial Distance (Normalized).} Normalize the distance from region centroid to edit box center by the edit box diagonal:

\begin{equation}
d_i^{\text{norm}} = \frac{\|\text{centroid}_i - \text{center}(B)\|_2}{\text{diag}(B)}
\end{equation}

where $\text{diag}(B) = \sqrt{(x_{\max} - x_{\min})^2 + (y_{\max} - y_{\min})^2}$. Normalization makes edit boxes of different sizes comparable.

\textbf{Dimension 2: Semantic Similarity.} Use CLIP (ViT-L/14)~\cite{radford2021learning} to compute cosine similarity between the edit region crop and spillover region crop:

\begin{equation}
s_i = \frac{\text{CLIP}(\text{crop}(I_{\text{gen}}, B)) \cdot \text{CLIP}(\text{crop}(I_{\text{gen}}, \text{bbox}_i))}{\|\text{CLIP}(\text{crop}(I_{\text{gen}}, B))\| \cdot \|\text{CLIP}(\text{crop}(I_{\text{gen}}, \text{bbox}_i))\|}
\end{equation}

When cropping, add 10-pixel padding to include context information. For efficiency, all region crops from the same image are batch-processed through CLIP inference at once, and the edit region feature is computed only once.

\textbf{Classification Rules.} Given distance threshold $\alpha$ and semantic threshold $\beta$:

\begin{equation}
\text{class}(R_i) = \begin{cases}
\text{spatial} & \text{if } d_i^{\text{norm}} < \alpha \text{ and } s_i \leq \beta \\
\text{semantic} & \text{if } d_i^{\text{norm}} \geq \alpha \text{ and } s_i > \beta \\
\text{mixed} & \text{if } d_i^{\text{norm}} < \alpha \text{ and } s_i > \beta \\
\text{random} & \text{if } d_i^{\text{norm}} \geq \alpha \text{ and } s_i \leq \beta
\end{cases}
\end{equation}

We set $\alpha = 1.5$ (i.e., 1.5 times the edit box diagonal as the near/far boundary) and $\beta = 0.80$.

\textbf{On the Choice of $\beta = 0.80$.} This threshold is significantly higher than commonly used CLIP similarity thresholds (typically 0.2-0.3). The reason lies in the specificity of text editing scenarios: cropped text regions visually share strong features (font rendering, text layout, background texture), resulting in a CLIP cosine similarity baseline of approximately 0.78 between any two text region crops (statistical mean from 5 models, 100K+ region pairs). If we used $\beta = 0.3$, 100\% of spillover regions would be judged as ``semantically related,'' completely losing discriminative power. $\beta = 0.80$ roughly corresponds to the median (P50) of the similarity distribution, effectively distinguishing ``genuine semantic associations above baseline'' from ``baseline-level visual similarity.'' We validate the robustness of this choice through ablation experiments in \cref{sec:ablation}.

\subsection{World Understanding Metrics}

We define two complementary metrics to quantify a model's implicit world knowledge:

\textbf{World Understanding Score (WUS)}: The ratio of semantic spillover to spatial spillover, measuring the ``quality'' of spillover:

\begin{equation}
\text{WUS} = \frac{|\{R_i : \text{class}(R_i) = \text{semantic}\}|}{|\{R_i : \text{class}(R_i) = \text{spatial}\}| + \epsilon}
\end{equation}

where $\epsilon = 0.01$ prevents division by zero. The intuitive meaning of WUS is: among the spillover produced by the model, how much is ``meaningful'' (semantically driven) vs. ``meaningless'' (attention leakage).

\textbf{Design Considerations for WUS}:
\begin{itemize}[leftmargin=*]
\item The numerator uses semantic rather than semantic + mixed, because the mixed type (close distance + semantically related) cannot exclude the contribution of attention leakage
\item The denominator uses spatial rather than total regions, to directly measure the ratio of ``understanding vs. leakage'' rather than being diluted by random types
\item When total spillover regions $< 5$, WUS is unstable and should be marked as N/A
\end{itemize}

\textbf{Limitations of WUS}: As a ratio metric, WUS can be inflated for low-spillover models. For example, an extremely conservative model might produce only 10 spillover regions (8 semantic, 2 spatial), yielding WUS = 4.0, but this does not mean it ``understands more'' than a model producing 100 spillover regions (60 semantic, 40 spatial, WUS = 1.5)—the latter activates more semantic associations.

\textbf{Semantic Spillover Density}: Average number of semantic spillover regions per image, measuring the ``quantity'' of spillover:

\begin{equation}
\text{SemanticDensity} = \frac{|\{R_i : \text{class}(R_i) = \text{semantic}\}|}{N_{\text{images}}}
\end{equation}

This metric directly reflects the model's absolute ability to activate semantic associations, unaffected by total spillover rate. High semantic density means the model frequently identifies and modifies semantically related content, demonstrating stronger world knowledge activation.

\section{EditSpilloverBench}
\label{sec:benchmark}

\subsection{Data Source}

We construct the benchmark from a production-level Chinese text editing quality audit dataset. The original data contains real-world scene images, each with bounding box annotations (marking text regions to be edited) and target replacement text.

\subsection{Dataset Composition}

EditSpilloverBench contains \textbf{200 images} covering 5 categories:

\begin{table}[t]
\centering
\small
\begin{tabular}{lcp{4.5cm}}
\toprule
\textbf{Category} & \textbf{Count} & \textbf{Description} \\
\midrule
App Screenshots & 54 & Mobile app interfaces with structured layout \\
Normal Docs & 23 & Documents, posters with mixed text/images \\
Others & 13 & Diverse scenarios \\
Real Scenes & 57 & Photos with signboards, road signs \\
Receipts & 53 & Invoices with high-density text \\
\bottomrule
\end{tabular}
\caption{EditSpilloverBench dataset composition (200 images)}
\label{tab:dataset}
\end{table}

\subsection{Controlled Test Groups}

To systematically evaluate spillover behavior, we design three controlled test groups corresponding to three quadrants of the 2$\times$2 classification matrix:

\begin{itemize}[leftmargin=*]
\item \textbf{Group A (Near + Unrelated)}: Text semantically unrelated to the edit target exists near the edit box. Expected to produce spatial spillover (attention leakage), should not produce semantic spillover.
\item \textbf{Group B (Far + Related)}: Text highly semantically related to the edit target exists far from the image (e.g., the same shop name appearing in different locations). If the model has world understanding, expected to produce semantic spillover.
\item \textbf{Group C (Far + Unrelated)}: Baseline control group, only unrelated content exists in the distance. Expected to produce no spillover or only a small amount of random spillover.
\end{itemize}

This design allows us to distinguish between attention leakage and world understanding mechanisms by comparing spillover patterns across Groups A/B/C.

\section{Experiments}
\label{sec:experiments}

\subsection{Experimental Setup}

\textbf{Evaluated Models.} We select 5 representative image editing models covering different architectures and capability levels:

\begin{table}[t]
\centering
\small
\begin{tabular}{lp{2.2cm}p{2.8cm}}
\toprule
\textbf{Model} & \textbf{Architecture} & \textbf{Rationale} \\
\midrule
Nano Banana & Production text editor & Spillover discovery source \\
Qwen-2511 & Dual-stream DiT & High quality representative \\
FireRed & Qwen-based & Architecture comparison \\
SeedReam 4.5 & Large pretrained & High spillover case \\
WanX 2.6 & Tongyi series & Medium spillover case \\
\bottomrule
\end{tabular}
\caption{Evaluated models}
\label{tab:models}
\end{table}

\textbf{Evaluation Metrics.} We report the following metrics:
\begin{itemize}[leftmargin=*]
\item \textbf{Spill\%} (spillover rate): proportion of modified pixels in non-edit region, lower is better
\item \textbf{SSIM} (structural similarity): degree of structure preservation in non-edit region, higher is better
\item \textbf{Spatial / Semantic / Mixed / Random}: proportion of region counts for the four spillover types
\item \textbf{WUS} (World Understanding Score): ratio of semantic spillover to spatial spillover, higher indicates more ``meaningful'' spillover
\end{itemize}

\textbf{Implementation Details.} Detection parameters: Gaussian blur $\sigma=2.0$, pixel threshold $\tau=15$, minimum region area $A_{\min}=100$. Classification parameters: $\alpha=1.5$, $\beta=0.80$, CLIP model ViT-L/14. Concurrency configuration: 8 threads, 8 GPUs (A100 80GB), batch CLIP inference (batch\_size=64). Full evaluation (5 models $\times$ ~200 images) takes approximately 24 minutes.

\subsection{Main Results}

Table~\ref{tab:main_results} shows the complete spillover analysis results for 5 models ($\beta=0.80$, $\alpha=1.5$).

\begin{table*}[t]
\centering
\small
\setlength{\tabcolsep}{3pt}
\begin{tabular}{lccccccccccc}
\toprule
\textbf{Model} & \textbf{Imgs} & \textbf{Spill\%} $\downarrow$ & \textbf{SSIM} $\uparrow$ & \textbf{Regions} & \textbf{Spat.} & \textbf{Sem.} & \textbf{Mix.} & \textbf{Rand.} & \textbf{WUS} & \textbf{Sem.Cnt} & \textbf{Sem.Den.} $\uparrow$ \\
\midrule
nano\_banana & 200 & 11.46 & 0.869 & 89.0 & 17.5 & 31.2 & 12.9 & 38.3 & 1.69 & 5561 & \textbf{27.8} \\
seedream45 & 198 & 11.12 & 0.864 & 240.8 & 21.1 & 26.7 & 7.0 & 45.1 & 1.21 & 12730 & \textbf{64.3} \\
firered & 200 & 7.55 & 0.901 & 79.6 & 12.9 & 36.4 & 15.0 & 35.7 & 2.62 & 5796 & \textbf{29.0} \\
wanx26 & 192 & 6.92 & 0.894 & 104.5 & 15.8 & 34.6 & 14.6 & 34.9 & 2.06 & 6950 & \textbf{36.2} \\
qwen\_2511 & 198 & 3.49 & 0.942 & 40.2 & 9.4 & 40.5 & 11.2 & 38.9 & 3.87 & 3222 & 16.3 \\
\bottomrule
\end{tabular}
\caption{Spillover analysis results. Spill\%: spillover rate (\%), Spat./Sem./Mix./Rand.: spillover type proportions (\%), Sem.Cnt: total semantic regions, Sem.Den.: semantic density (regions/image).}
\label{tab:main_results}
\end{table*}

\begin{figure}[t]
\centering
\includegraphics[width=0.9\columnwidth]{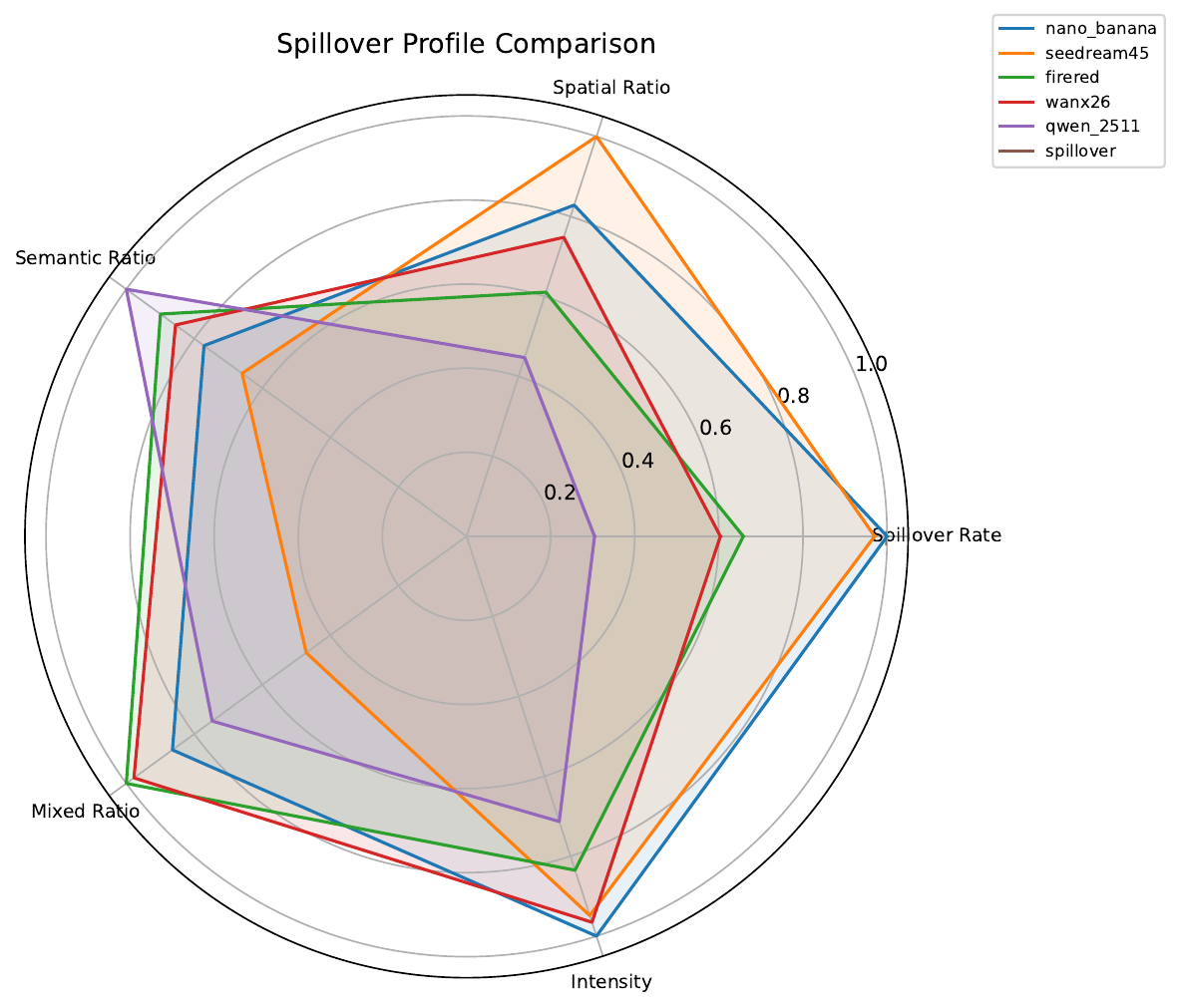}
\caption{Cross-model comparison radar chart. Models exhibit different trade-offs between spillover control (SSIM, low Spill\%) and world understanding (WUS, Semantic Density).}
\label{fig:radar}
\end{figure}

From \cref{tab:main_results,fig:radar}, we extract the following key observations:

\textbf{Observation 1: Spillover rates vary significantly across models.} Spillover rates range from 3.49\% (qwen\_2511) to 11.46\% (nano\_banana), with a 3.3$\times$ ratio between highest and lowest. This difference far exceeds random fluctuation, reflecting fundamental differences in edit region constraint capabilities across model architectures. Notably, while seedream45's spillover rate (11.12\%) is close to nano\_banana's, its average number of spillover regions is 2.7 times higher (240.8 vs. 89.0), and total spillover area is 3.8 times higher (438,224 vs. 116,197 pixels), indicating that seedream45's spillover is more diffuse and fragmented, while nano\_banana's spillover is more concentrated.

\textbf{Observation 2: Semantic spillover density reveals differences in world understanding capability.} This is one of the most important findings of this paper. Sorted by semantic spillover density: seedream45 (64.3) $>$ wanx26 (36.2) $>$ firered (29.0) $>$ nano\_banana (27.8) $>$ qwen\_2511 (16.3). However, considering that seedream45's spillover is extremely fragmented (240.8 regions/image) and has the highest proportion of random spillover (45.1\%), its high density mainly stems from noise rather than genuine semantic understanding.

Excluding seedream45, \textbf{nano\_banana demonstrates the strongest world understanding activation capability}:
\begin{itemize}[leftmargin=*]
\item Semantic spillover density 27.8, highest among non-fragmented models
\item Semantic spillover proportion 31.2\%, absolute count 5561
\item More concentrated spillover (89.0 regions/image vs. seedream45's 240.8), indicating targeted semantic activation
\end{itemize}

In contrast, while qwen\_2511 has the highest WUS (3.87), this is mainly because its total spillover is minimal (3.49\%), causing the ratio to be inflated. Its semantic spillover density is only 16.3, less than 60\% of nano\_banana's, indicating it is most precise in editing control but activates fewer semantic associations.

\textbf{Observation 3: Trade-off between editing control and world understanding activation.} qwen\_2511 has the highest SSIM (0.942) and lowest spillover rate (3.49\%), making it the most precise editing control model. But its semantic spillover density is lowest (16.3), indicating it tends toward a conservative strategy—strictly limiting the edit region and reducing semantic association activation. nano\_banana exhibits opposite characteristics: higher spillover rate (11.46\%) accompanied by strongest semantic activation (27.8 density), indicating it more actively identifies and modifies semantically related content. This trade-off reflects differences in model training objectives: qwen\_2511 optimizes content preservation, nano\_banana optimizes semantic consistency.

\subsection{Spatial Decay Analysis}

To distinguish between attention leakage and world understanding spillover mechanisms, we analyze how spillover intensity decays with distance from the edit box.

\textbf{Method.} Divide normalized distance $d^{\text{norm}}$ into 7 bins: [0, 0.5), [0.5, 1.0), [1.0, 1.5), [1.5, 2.0), [2.0, 3.0), [3.0, 5.0), [5.0, 10.0). For each bin, compute \textbf{spillover area density}: total area of all spillover regions in that bin divided by the annular area corresponding to that distance bin ($\pi(r_{\text{outer}}^2 - r_{\text{inner}}^2)$), to eliminate bias from larger areas at greater distances.

\begin{figure}[t]
\centering
\includegraphics[width=\columnwidth]{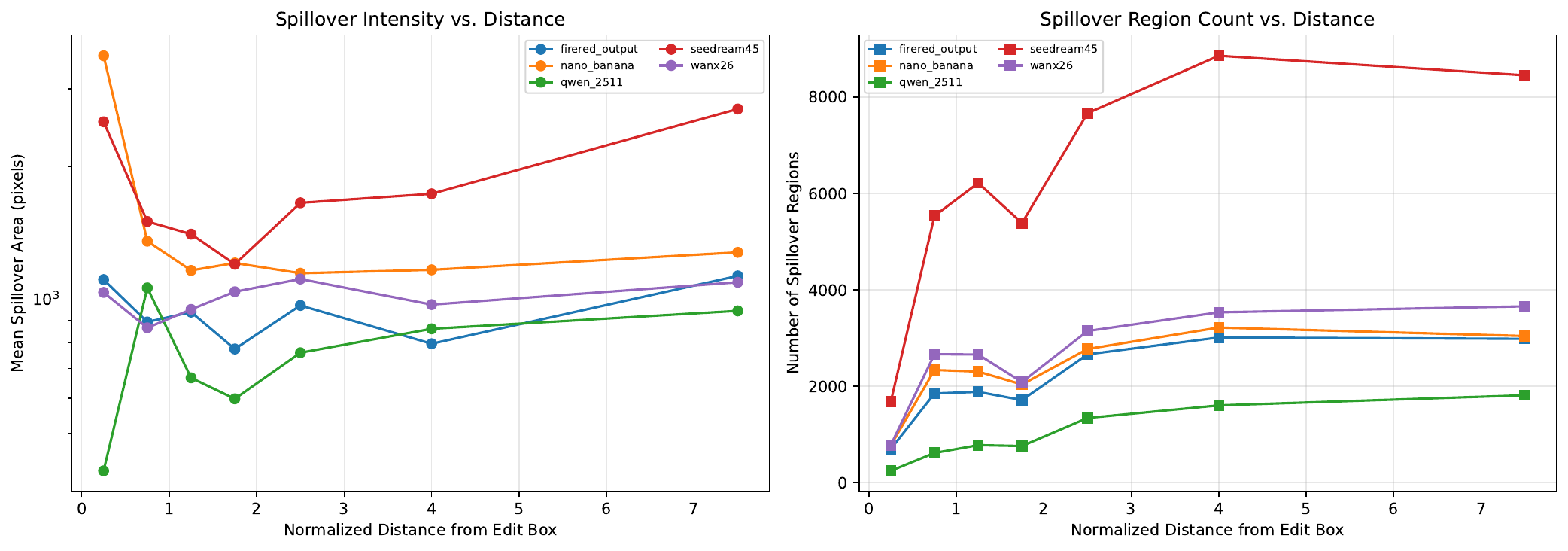}
\caption{Spillover area density decay with distance. All models exhibit exponential decay, with nano\_banana showing the steepest decline (200$\times$ from near to far distance).}
\label{fig:spatial_decay}
\end{figure}

\begin{table*}[t]
\centering
\small
\setlength{\tabcolsep}{4pt}
\begin{tabular}{lccccc}
\toprule
\textbf{Distance} & \textbf{nano\_banana} & \textbf{seedream45} & \textbf{firered} & \textbf{wanx26} & \textbf{qwen\_2511} \\
\midrule
0--0.5$\times$ & 100 & 100 & 100 & 100 & 100 \\
0.5--1.0$\times$ & 38.4 & 65.4 & 70.9 & 94.8 & 216.7 \\
1.0--1.5$\times$ & 19.5 & 41.3 & 45.6 & 62.4 & 102.8 \\
1.5--2.0$\times$ & 12.8 & 21.8 & 24.5 & 38.4 & 64.3 \\
2.0--3.0$\times$ & 5.8 & 15.0 & 16.7 & 21.6 & 50.5 \\
3.0--5.0$\times$ & 2.1 & 5.7 & 4.8 & 6.6 & 21.3 \\
5.0--10.0$\times$ & 0.5 & 1.8 & 1.5 & 1.6 & 5.7 \\
\bottomrule
\end{tabular}
\caption{Spillover area density decay with distance (normalized to near distance = 100\%). Distance is measured in multiples of edit box diagonal.}
\label{tab:spatial_decay}
\end{table*}

\textbf{Finding 1: Most models exhibit approximately exponential decay.} nano\_banana's decay is the steepest: from near distance to 2-3$\times$ distance it drops to 5.8\%, and at 5-10$\times$ distance only 0.5\% remains, a decay of approximately 200$\times$ (see \cref{fig:spatial_decay}). This exponential decay pattern is consistent with the spatial locality of attention mechanisms—attention weights decrease rapidly with distance, causing spatial diffusion of the edit signal to decay exponentially.

\textbf{Finding 2: Decay rates vary by model.} nano\_banana decays fastest, seedream45 and firered are intermediate, wanx26 is slower, and qwen\_2511 is slowest. Notably, qwen\_2511's density at 0.5-1.0$\times$ distance is actually \emph{higher} than at near distance (216.7\%). This anomaly may be because qwen\_2511's near-distance spillover is effectively suppressed (total spillover rate only 3.49\%), making the small amount of mid-distance spillover appear relatively prominent after normalization.

\textbf{Finding 3: Semantic spillover proportion is distance-invariant (key evidence).} We further analyze the proportion of semantically relevant spillover ($s_i > \beta$) within each distance bin (\cref{tab:semantic_proportion,fig:semantic_dist}):

\begin{figure}[t]
\centering
\includegraphics[width=\columnwidth]{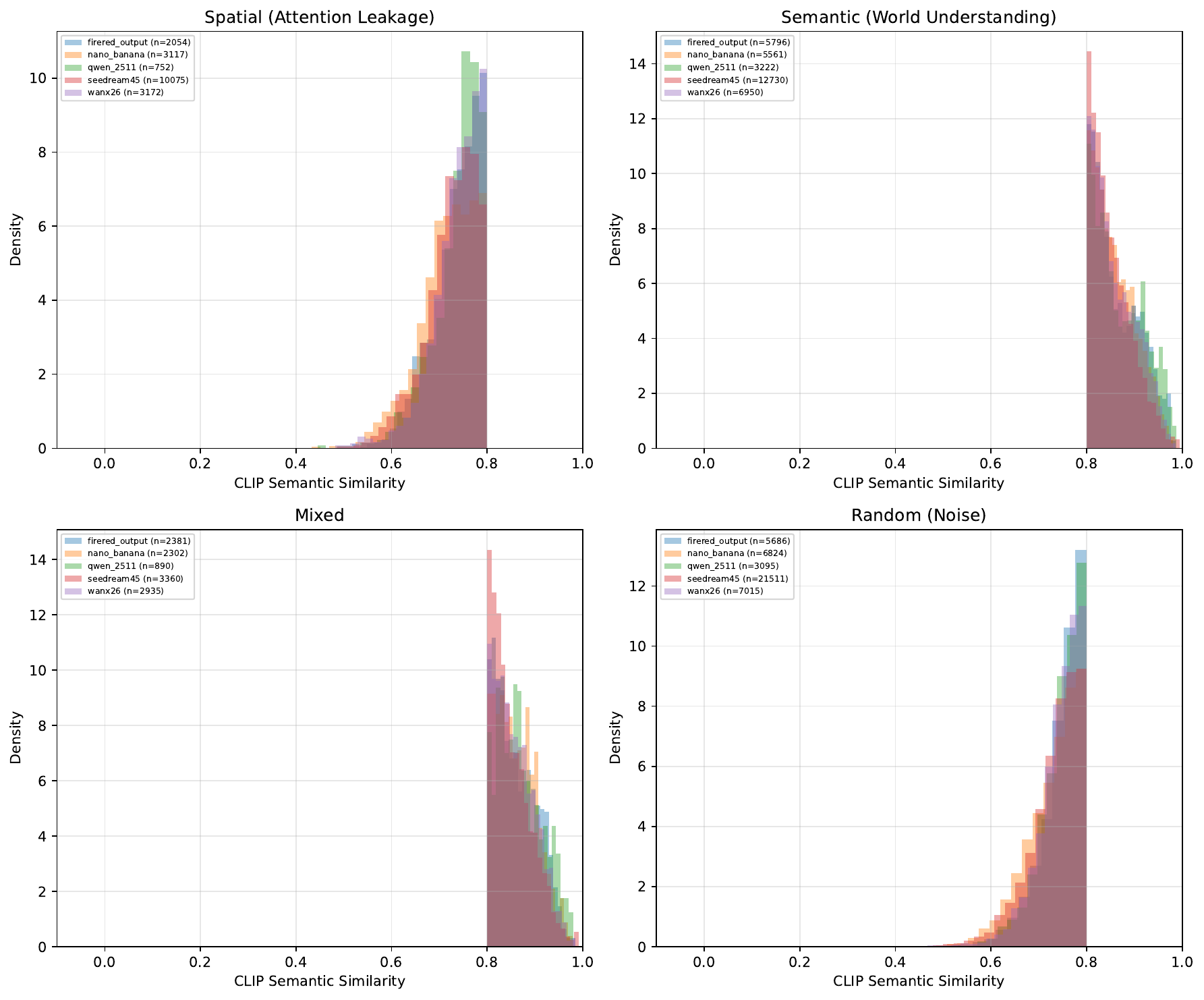}
\caption{Semantic spillover proportion remains constant across distances (39\%--58\%), providing direct evidence that semantic spillover is not driven by spatial proximity.}
\label{fig:semantic_dist}
\end{figure}

\begin{table}[t]
\centering
\small
\begin{tabular}{lcc}
\toprule
\textbf{Distance} & \textbf{nano\_banana} & \textbf{qwen\_2511} \\
\midrule
0--0.5$\times$ & 53.2 & 58.5 \\
0.5--1.0$\times$ & 42.4 & 56.4 \\
1.0--1.5$\times$ & 39.0 & 51.1 \\
1.5--2.0$\times$ & 38.7 & 50.9 \\
2.0--3.0$\times$ & 46.5 & 52.5 \\
3.0--5.0$\times$ & 41.7 & 48.6 \\
5.0--10.0$\times$ & 48.6 & 49.3 \\
\bottomrule
\end{tabular}
\caption{Semantic spillover proportion (\%) by distance}
\label{tab:semantic_proportion}
\end{table}

nano\_banana's semantic proportion fluctuates between 39\%--53\%, qwen\_2511 between 49\%--58\%, both without clear distance-dependent trends. The implication of this result is profound:

\begin{itemize}[leftmargin=*]
\item \textbf{Attention leakage} (spatial spillover) decays exponentially with distance—this is a physical property of the architecture
\item The proportion of \textbf{semantic spillover} remains constant across distance bins—this means semantic spillover is not driven by spatial proximity
\end{itemize}

The decoupling of these two provides direct evidence: semantic spillover reflects internal semantic association activation in the model, not spatial diffusion of attention. When a model modifies semantically related content in the distance, it is not because ``attention leaked there,'' but because it ``understood the relationship between that content and the edit target.''

\subsection{Quality-Spillover Trade-off}

Table~\ref{tab:main_results} reveals a counterintuitive pattern: spillover rate and WUS exhibit an inverse relationship. We conduct a deeper analysis of this.

\textbf{Spillover ``quality'' matters more than ``quantity.''} qwen\_2511 has the lowest spillover rate (3.49\%), but when it does produce spillover, 40.5\% is semantic (far distance + semantically related), and only 9.4\% is spatial (near distance + semantically unrelated). In contrast, seedream45 has the highest spillover rate (11.12\%), but 21.1\% of its spillover is spatial, 26.7\% is semantic, and 45.1\% is random—much of the spillover is ``meaningless.''

\textbf{Mechanistic explanation of this trade-off.} We hypothesize that models with stronger editing control (such as qwen\_2511) learned more precise edit region constraints during training, effectively suppressing spatial leakage of attention. But semantic associations are deeper features—they are encoded in the model's semantic representations and are not easily eliminated by simple spatial constraints. Therefore, models with better control ``filter out'' spatial leakage but ``preserve'' semantic associations, resulting in higher WUS.

\textbf{Practical implications.} This finding has direct guidance for model development: if the goal is to reduce spillover, focus should be on suppressing spatial leakage (e.g., improving spatial locality of attention), rather than indiscriminately suppressing all changes in non-edit regions—the latter may simultaneously suppress valuable semantic consistency.

\section{Ablation Study}
\label{sec:ablation}

Our classification framework relies on two key hyperparameters: semantic threshold $\beta$ (CLIP cosine similarity threshold) and distance threshold factor $\alpha$ (spatial distance threshold = $\alpha \times$ edit box diagonal). This section validates the robustness of our core conclusions to these two parameters.

\subsection{Impact of Semantic Threshold $\beta$}

We test across $\beta \in \{0.70, 0.75, 0.80, 0.85, 0.90\}$ with fixed $\alpha = 1.5$.

\begin{table}[t]
\centering
\small
\begin{tabular}{lccccc}
\toprule
$\beta$ & \textbf{nano} & \textbf{seed} & \textbf{fire} & \textbf{wanx} & \textbf{qwen} \\
\midrule
0.70 & 8.38 & 9.13 & 17.09 & 15.72 & 25.67 \\
0.75 & 3.81 & 3.35 & 7.16 & 5.71 & 10.79 \\
0.80 & 1.69 & 1.21 & 2.62 & 2.06 & 3.87 \\
0.85 & 0.66 & 0.44 & 0.88 & 0.71 & 1.49 \\
0.90 & 0.21 & 0.13 & 0.37 & 0.25 & 0.66 \\
\bottomrule
\end{tabular}
\caption{WUS values under different $\beta$ ($\alpha = 1.5$)}
\label{tab:ablation_beta}
\end{table}

\textbf{WUS ranking is completely consistent across all $\beta$ values}: qwen\_2511 $>$ firered $>$ wanx26 $>$ nano\_banana $>$ seedream45. This stability indicates that our core finding—the inverse relationship between spillover rate and WUS—does not depend on specific threshold choices.

\textbf{Rationale for $\beta$ selection.} We choose $\beta = 0.80$ as the default value for the following reasons: (1) In text editing scenarios, CLIP's baseline similarity for text crops is high (mean 0.78--0.81); too low a threshold (e.g., 0.30) would classify almost all regions as ``semantically related,'' losing discriminative power; (2) $\beta = 0.80$ is approximately at the median (P50) of the CLIP similarity distribution, effectively distinguishing high-similarity from low-similarity regions; (3) At this threshold, the distribution of the four spillover types is relatively balanced, avoiding extreme skew.

\begin{table}[t]
\centering
\small
\begin{tabular}{lcccc}
\toprule
\textbf{Model} & \textbf{Spat.} & \textbf{Sem.} & \textbf{Mix.} & \textbf{Rand.} \\
\midrule
nano\_banana & 17.5 & 31.2 & 12.9 & 38.3 \\
seedream45 & 21.1 & 26.7 & 7.0 & 45.1 \\
firered & 12.9 & 36.4 & 15.0 & 35.7 \\
wanx26 & 15.8 & 34.6 & 14.6 & 34.9 \\
qwen\_2511 & 9.4 & 40.5 & 11.2 & 38.9 \\
\bottomrule
\end{tabular}
\caption{Classification distribution (\%) at $\beta = 0.80$}
\label{tab:classification_dist}
\end{table}

\subsection{Impact of Distance Threshold Factor $\alpha$}

We test across $\alpha \in \{1.0, 1.5, 2.0\}$ with fixed $\beta = 0.80$.

\begin{table}[t]
\centering
\small
\begin{tabular}{lccccc}
\toprule
$\alpha$ & \textbf{nano} & \textbf{seed} & \textbf{fire} & \textbf{wanx} & \textbf{qwen} \\
\midrule
1.0 & 3.42 & 2.45 & 5.36 & 4.36 & 8.03 \\
1.5 & 1.69 & 1.21 & 2.62 & 2.06 & 3.87 \\
2.0 & 1.05 & 0.78 & 1.60 & 1.33 & 2.35 \\
\bottomrule
\end{tabular}
\caption{WUS values under different $\alpha$ ($\beta = 0.80$)}
\label{tab:ablation_alpha}
\end{table}

WUS ranking is also completely consistent across all $\alpha$ values. Changes in $\alpha$ mainly affect the absolute value of WUS (smaller $\alpha$ means narrower ``near distance'' range, more regions classified as ``far distance,'' overall WUS increases), but do not change the relative ordering between models.

\subsection{Summary}

Ablation experiments show that our core findings—qwen\_2511 has the highest WUS, seedream45 the lowest, and spillover rate and WUS exhibit an inverse relationship—remain stable across a wide range of $\beta \in [0.70, 0.90]$ and $\alpha \in [1.0, 2.0]$. This robustness strengthens our confidence in the conclusions: differences in world understanding capabilities between models are real, not artifacts of threshold selection.

\section{Discussion}
\label{sec:discussion}

\subsection{Limitations}

\textbf{Scenario Coverage.} The current EditSpilloverBench only covers Chinese text editing scenarios (5 categories: app screenshots, normal scenes, receipts, real photos, others). The spillover taxonomy we proposed in \cref{sec:benchmark} encompasses dimensions such as physical consistency (lighting/shadows), causal understanding (seasonal/temporal changes), and cultural common sense (brand/regional styles), but these dimensions have not yet been validated in experiments. Extending the framework to these scenarios requires constructing corresponding test datasets, which is important future work.

\textbf{Proxy Indicator for Semantic Relevance.} We use CLIP cosine similarity as a measure of ``semantic relevance.'' While CLIP excels at vision-language alignment, it captures visual appearance-level similarity and may not fully reflect deep semantic relationships (such as causal or functional relationships). For example, the similarity between ``signboard'' and ``menu'' in CLIP space may be lower than between ``signboard'' and ``another signboard,'' even though the former semantic association is more meaningful. Future work could explore using multimodal large language models (such as GPT-4V) for more fine-grained semantic relationship judgments.

\textbf{Limitations of Pixel-level Detection.} Our detection pipeline is based on pixel-level differences and is sensitive to JPEG compression artifacts. Although we perform denoising through Gaussian blur ($\sigma = 2.0$) and minimum area filtering (100 pixels), false positives may still occur on highly compressed images. Additionally, pixel-level differences cannot capture semantic-level changes (such as text replacements where color remains unchanged but meaning changes), which require higher-level detection methods such as OCR.

\textbf{Model Coverage.} We evaluated 5 representative models covering different architectures and training strategies. However, image editing models are developing rapidly, and new models (such as DiT architecture-based editing models) may exhibit different spillover patterns. Our framework is designed to be model-agnostic and can be directly applied to evaluation of new models.

\subsection{Mechanistic Explanation of Semantic Spillover}

The semantic spillover phenomenon we observe—models modifying semantically related but unspecified content in the distance—may originate from the interaction between cross-attention and self-attention in diffusion models.

In U-Net or DiT-based diffusion models, text conditions are injected into image features through cross-attention~\cite{hertz2022prompt,mokady2023null}. When an editing instruction involves a specific semantic concept (such as ``shop name''), cross-attention may activate all regions in the image related to that concept—not only the target region inside the edit box, but also semantically related regions in the distance (such as related text on a menu). Subsequently, self-attention propagates information between these activated regions, causing distant regions to also be modified.

This mechanistic explanation accounts for why semantic spillover is distance-invariant: cross-attention activation is based on semantic matching rather than spatial proximity, so semantically related regions may be activated regardless of distance. In contrast, spatial spillover is primarily driven by the spatial locality of self-attention, and therefore decays exponentially with distance.

\subsection{Implications for Model Development}

Our findings have the following implications for the development and evaluation of image editing models:

\textbf{Spillover should not be indiscriminately suppressed.} Traditional evaluation treats all changes in non-edit regions as ``errors.'' Our analysis shows that some spillover (semantic type) actually reflects the model's understanding of world relationships and is a valuable capability signal. Future evaluation frameworks should distinguish between ``harmful spillover'' (spatial leakage) and ``beneficial spillover'' (semantic consistency).

\textbf{Editing control and world understanding can coexist.} The case of qwen\_2511 shows that low spillover rate and high WUS can be achieved simultaneously. This suggests an ideal state of ``precise editing'' exists: the model strictly controls the edit region but can make semantically consistent global adjustments when necessary.

\textbf{WUS can serve as a complementary metric for model selection.} Among models with similar editing quality, WUS provides an additional evaluation dimension: higher WUS means the model has stronger implicit world understanding capabilities and may perform better on editing tasks requiring global consistency.

\section{Conclusion}
\label{sec:conclusion}

We propose EditSpilloverProbe, a systematic framework that repurposes the ``edit spillover'' phenomenon in image editing models as a probe for world knowledge. By introducing a 2$\times$2 spillover taxonomy (spatial distance $\times$ semantic relevance) and two complementary metrics (WUS and semantic spillover density), we transform spillover from a ``bug'' that needs to be eliminated into a diagnostic signal that can be quantitatively analyzed.

Systematic evaluation of 5 representative editing models yields three core findings:

\begin{enumerate}[leftmargin=*]
\item \textbf{Spillover rates vary significantly across models} (3.49\%--11.46\%, 3.3$\times$ difference), and are closely related to model architecture and training strategies.

\item \textbf{Trade-off between editing control and world understanding activation}—qwen\_2511 demonstrates the strongest editing control (spillover rate 3.49\%, SSIM 0.942) but lowest semantic spillover density (16.3/image); nano\_banana, as a production-grade model, demonstrates the strongest world understanding activation capability (semantic spillover density 27.8/image, absolute count 5561), but with higher total spillover rate (11.46\%). This trade-off reveals the opposition between two optimization objectives: content preservation vs. semantic consistency. Our analysis shows that \textbf{absolute semantic spillover quantity better reflects models' world knowledge activation capability than WUS ratio}.

\item \textbf{Semantic spillover proportion is distance-invariant}—although total spillover area decays exponentially with distance (consistent with attention leakage), the proportion of semantically relevant spillover remains constant across distance bins (40\%--58\%), providing direct evidence that semantic spillover reflects genuine world understanding rather than spatial diffusion.
\end{enumerate}

Our methodology—using models' unintended behavior as a passive probe—provides a new paradigm for capability evaluation of generative models. Unlike traditional task-design-based evaluation, behavior-level probing does not require special test cases, is applicable to black-box models, and can reveal implicit capabilities of models during natural use. In the future, we plan to extend this framework to broader world knowledge dimensions such as physical consistency and causal understanding, and explore its applications in domains such as video editing and 3D generation.

\bibliographystyle{ieee_fullname}
\bibliography{references}

\end{document}